\documentclass[10pt,twocolumn,letterpaper]{article}

\usepackage{iccv}
\usepackage{times}
\usepackage{epsfig}
\usepackage{graphicx}
\usepackage{amsmath}
\usepackage{amssymb}

\usepackage{multicol}
\usepackage{multirow}
\usepackage{array}
\usepackage{booktabs}
\usepackage{graphicx}
\usepackage{tikz}
\usepackage{subcaption}
\usepackage[numbers,sort,compress]{natbib}

\usepackage[pagebackref=true,breaklinks=true,letterpaper=true,colorlinks,bookmarks=false]{hyperref}

\iccvfinalcopy 

\def\iccvPaperID{30} 

\begin{document}

\title{An empirical study of the effect of video encoders on Temporal Video Grounding}

\author{Ignacio M. De la Jara\text{\small$^{\dag}$} \quad Cristian Rodriguez-Opazo\text{\small$^{\ddagger}$} \quad Edison Marrese-Taylor\text{\small$^{\clubsuit}$} \quad Felipe Bravo-Marquez\text{\small$^{\dag}$} \\
{\large Department of Computer Science, University of Chile, CENIA and IMFD \text{\small$^\dag$}}\\
{\large Australian Institute for Machine Learning, University of Adelaide\text{\small$^\ddagger$}}\\
{\large National Institute of Advanced Industrial Science and Technology \text{\small$^\clubsuit$}}\\
{\tt\small ignacio.meza@ug.uchile.cl, cristian.rodriguezopazo@adelaide.edu.au} \\
{\tt\small emarrese@weblab.t.u-tokyo.ac.jp, fbravo@dcc.uchile.cl}
}

\maketitle %

\begin{abstract}
    Temporal video grounding is a fundamental task in computer vision, aiming to localize a natural language query in a long, untrimmed video. It has a key role in the scientific community, in part due to the large amount of video generated every day. Although we find extensive work in this task, we note that research remains focused on a small selection of video representations, which may lead to architectural overfitting in the long run. To address this issue, we propose an empirical study to investigate the impact of different video features on a classical architecture. 
    We extract features for three well-known benchmarks, Charades-STA, ActivityNet-Captions and YouCookII, using video encoders based on CNNs, temporal reasoning and transformers. 
    Our results show significant differences in the performance of our model by simply changing the video encoder, while also revealing clear patterns and errors derived from the use of certain features, ultimately indicating potential feature complementarity.
    
\end{abstract}

\vspace{-.5cm}
\section{Introduction}

Understanding and reasoning about long, untrimmed videos is at the core of Computer Vision (CV). As humans have the ability to intuitively identify relevant moments within videos, the task of Temporal Video Grounding (TVG) appears as a fundamental effort in CV, aiming to develop models that recognise and determine temporal boundaries of action instances in videos \cite{Shou_2016_CVPR,guAVAVideoDataset2018,girdharVideoActionTransformer2019} using natural language queries \cite{Gao_2017_ICCV,Hendricks_2017_ICCV}.



As such, work on the TVG task is extensive and includes a wide variety of approaches and techniques. While the original models were mostly suggestion-based, more recent techniques have aimed at predicting the start and end temporal positions directly, or by regressing them from the input video. The recent advent of transformer-based models \cite{vaswaniAttentionAllYou2017} has also brought new developments, where the integration of pre-training stages or the direct addition of pre-trained vision and speech models has led to significant performance improvements.


Despite the large amount of prior work in the TVG task, we find that the role of the video representation has not been consistently investigated so far. Specifically, we observe that prior work has relied on features derived from action classification models such as C3D \cite{tranLearningSpatiotemporalFeatures2015a} or I3D \cite{carreiraQuoVadisAction2017}, with alternatives remaining relatively unexplored. We speculate that this selection bias may lead to model designs that overfit or exploit spurious patterns in these features, without generalising in the long run. We suggest that the increased complexity of recent approaches in search of improved performance can be seen as indirect evidence of this point.



In light of this issue, in this paper we propose a comprehensive empirical study to shed light on the role of video representation in the TVG task. We consider three relevant benchmark datasets, Charades-STA, ActivityNet Captions, and YoucookII, and design a comprehensive framework that allows us to isolate the effect of different video representations by introducing minimal changes in the model architecture. We use a wide variety of pre-trained models to extract video representations, including more than 10 different types of models, resulting in more than 30 sets of extracted features for our data. These include, but are not limited to, well-known CNN and Transformer-based action classifiers. We release features to encourage research in this area. \footnote{\url{https://github.com/Mezosky/VideoFeatures_TVG}}

Our results show that changes in the video representation can lead to substantial performance improvements by keeping the models as is, allowing a ``classical'' approach \cite{rodriguezProposalfreeTemporalMoment2020} to perform better by a large margin by simply changing the video encoder, findings that are consistent with similar observations recently made in the context of query representations \cite{kido-shimomoto-etal-2023-towards}. 
Our experiments reveal clear pitfalls in the use of certain features and uncover complementarity between features, which could lead to further performance improvements if exploited. 

\section{Related Work}

\paragraph{Action Classification} The task of identifying the action being performed in a video is the cornerstone of current video understanding pipelines. It is essential for the task of temporal video grounding, as it is the current way of encoding videos. Current methods are usually trained via supervised learning, using datasets such as Something-Something \cite{goyal2017something}, Kinetics \cite{carreiraQuoVadisAction2017}, among others, which contain short video clips of a single action.

Different methods have been proposed to solve the action classification task, the most recent approach based on neural networks could be divided into CNN-based \cite{Tran_2018_CVPR,carreiraQuoVadisAction2017, feichtenhoferSlowFastNetworksVideo2019,feichtenhofer2020x3d}, temporal reasoning \cite{linTSMTemporalShift2019,limin2019temporalsegment} and transformers \cite{fanMultiscaleVisionTransformers2021,liMViTv2ImprovedMultiscale2022}. In the case of CNN-based, the methods exploit the ability of CNNs to capture spatial information in images and add various techniques to extract the temporal information from the video. \cite{Tran_2018_CVPR} proposed a convolutional block, which is the combination of 2D convolutions followed by a 1D convolution. Such blocks approximate the behaviour of 3D convolutions and allow to capture the temporality from the videos. Knowing that the information in a video is very redundant and not symmetric, \cite{feichtenhoferSlowFastNetworksVideo2019} proposed to use a slow and a fast track, aiming to capture the spatial and temporal information respectively.
Temporal reasoning classifiers aim to capture the temporal information between features extracted from frames using techniques that exploit latent information in videos, such as redundancy (sampling), summarisation (aggregation), and ordering (ranking). Limin et al. \cite{limin2019temporalsegment} use a sampling technique to select features that, after passing through an aggregation mechanism, effectively represent the video. Aiming to create a better representation of the videos, Lin et al. \cite{linTSMTemporalShift2019} proposed a shift module that applies a shift in part of the temporal channel of the features to exchange information between frames.
The advent of transformer-based models has also led to new developments in this task \cite{fanMultiscaleVisionTransformers2021, carionEndtoEndObjectDetection2020}. However, their success in this task is due to their flexibility in capturing spatial information and their inherent ability to model sequential information. \cite{fanMultiscaleVisionTransformers2021} presents a multiscale feature hierarchy that combines low-level spatial information processed at early layers with more complex high-level features that capture temporal information processed at deeper layers.
\vspace{-0.3cm}
\paragraph{Temporal Video Grounding} The task that concerns this work, where we find mainly two types of approaches. On the one hand, we find proposal-based models that, given a query, return a set of candidates that can later be ranked \cite{liu2018attentive,ge2019mac,sap2019,xu2019multilevel,zhangMultiScale2DTemporal2021}. Recent approaches include both transformer-based models and contrastive losses \cite{rodriguez2023memory,rodriguez2021locformer,wangNegativeSampleMatters2022,zhangMultiStageAggregatedTransformer2021,zhengWeaklySupervisedTemporal2022}. On the other hand, we find proposal-free models that try to predict the start and end locations directly from the video span \cite{ghosh_excl_2019,rodriguezProposalfreeTemporalMoment2020}.



\section{Experimental Framework}

\paragraph{Model} At the core of our experimental framework lies the TVG model to be used as a pivot for testing video features. Although there are a variety of approaches to our task, we find that not all of them are suitable for our purposes, as replacing the video representations can in many cases lead to substantial changes in the model architecture and/or training. After careful consideration, we selected the proposal-free approach proposed by Rodriguez et al. \cite{rodriguezProposalfreeTemporalMoment2020} (TMLGA), which can be summarized in three parts: 1) a text encoder based on GloVe \cite{pennington2014glove} embeddings on top of an LSTM \cite{hochreiter1997long}, 2) a video encoder, which we discuss in detail below, and, 3) a localization module that combines information from both modalities using a dynamic filter \cite{jia2016dynamic} and predicts the start and end points of the segment. We consider this model to be a ``\textit{classic}'' approach, as it has been widely adopted as a baseline by recent work, allowing for meaningful comparisons with a wide variety of approaches. We also note that TMLGA relies on a rather simple approach to query representation, which we believe allows for a cleaner result in terms of the contribution of the language encoder. Finally, we point out that TMLGA has an official implementation that has been publicly released, which we believe is essential to facilitate the replication of the original results and ultimately ensure the reproducibility of our experiments.
\vspace{-.4cm}
\paragraph{Video Representation} Following previous work, our pivot TMLGA model uses pre-trained action classification models to obtain a sequence of vectors to represent a given video. Critically, these input vectors are just passed through one projection layer before being fed into the localization module, which ensures that the amount of changes introduced to the model is minimized for our experiments. Although the original implementation relies on I3D \cite{carreiraQuoVadisAction2017} we instead consider a wide selection of pre-trained models for feature extraction, which we propose to group into three main categories. First, we test CNN-based video classification approaches, including C2D \cite{tranLearningSpatiotemporalFeatures2015a}, I3D, SlowFast \cite{feichtenhoferSlowFastNetworksVideo2019}, X3D \cite{feichtenhofer2020x3d} and Non-Local variations in one classifier \cite{wang2018nonlocal}. For these models, we modify the classification head of the last ResNet or ConvNet used, extracting the representation generated before dropout, final projections and classification layer of the network. Secondly, we look at Transformer-based action classification models, including  MViT \cite{fanMultiscaleVisionTransformers2021}, MViT2 \cite{liMViTv2ImprovedMultiscale2022}, and Rev-MViT \cite{mangalam2023reversible}. Similarly to thee CNN-based case, for these models we represent the video using the vectors of the last layer of the transformer before the linear classifier. Finally, we study temporal reasoning models such as TSM \cite{linTSMTemporalShift2019}, which provide a different angle in tackling the action classification task. In this model, the video data is divided into segments, each containing 8 buckets, where each bucket consists of 8 consecutive frames. This segmentation approach creates smaller video subsequences, allowing the temporal network to effectively permute frames and capture their temporal dependencies. Similarly to the other cases, the last projections of the temporal network are discarded to obtain the video feature representation.

Following Rodriguez et al. \cite{rodriguezProposalfreeTemporalMoment2020}, the video encoding procedure is done offline, performing rescaling to 320x240 when necessary, and sampling at 25 fps. Our implementation is based on \textit{decord}\footnote{\url{https://github.com/dmlc/decord}}, \textit{pyslowfast}\footnote{\url{https://github.com/facebookresearch/SlowFast}} and the official code release for TSM \cite{linTSMTemporalShift2019} \footnote{\url{https://github.com/mit-han-lab/temporal-shift-module\#pretrained-models}}. Each video is sequentially traversed in buckets of a size given by the frame rate of the backbone model. In cases where the total number of extracted frames is not divisible by this number, we fill the last bucket by repeating the last frame.

\vspace{-.3cm}
\paragraph{Datasets} We consider three well-known challenging benchmarks for the task. \textit{Charades-STA} \cite{sigurdssonHollywoodHomesCrowdsourcing2016a, Gao_2017_ICCV}, \textit{ActivityNet-Captions} \cite{krishnaDenseCaptioningEventsVideos2017}, and \textit{YouCookII} \cite{zhouAutomaticLearningProcedures2018,zhouWeaklySupervisedVideoObject2018}. We use the pre-defined train and test sets. Each dataset with its own challenges, from more ambiguous spatio-temporal information to very long videos.
\vspace{-.3cm}
\paragraph{Evaluation} To assess the effectiveness of our extracted features, we conduct two kinds of analysis. First, a quantitative one based on metrics proposed in \cite{Gao_2017_ICCV}, namely the recall with thresholds ($\alpha$) of 0.5 and 0.7 for the temporal intersection over union (tIoU). Secondly, we provide an in-depth qualitative analysis of the model predictions, focusing on the biases presented when training with different features.



\section{Results}

\begin{table}[t]
    \centering
    \resizebox{0.9\columnwidth}{!}{%
    \scriptsize
    \begin{tabular}{l@{\hspace{0.15cm}} c@{\hspace{0.15cm}} c@{\hspace{0.15cm}} c@{\hspace{0.15cm}} c@{\hspace{0.15cm}} c@{\hspace{0.15cm}} c@{\hspace{0.15cm}} c@{\hspace{0.15cm}} c@{\hspace{0.15cm}} }
        \toprule
        \multicolumn{3}{c}{\textbf{Model}} & \multicolumn{2}{c}{\bf Charades} & \multicolumn{2}{c}{\bf ActivityNet} & \multicolumn{2}{c}{\bf YouCookII} \\
        Encoder & $f$ & Pretrain & 0.5 & 0.7 & 0.5 & 0.7 & 0.5 & 0.7 \\
        \midrule
        MViT     & 8  & K400   & \textbf{50.27} & 31.32 & 30.64 & 17.50 & \textbf{27.78} & \textbf{15.21}    \\
        MViT-v2  & 16 & K400   & 48.17 & 30.27 & 23.80 & 11.87 & 25.40 & 13.95    \\
        Rev-MViT & 16 & K400   & 49.89 & \textbf{32.31} & 28.94 & 15.97 & 23.71 & 13.00    \\
        X3D M    & 16 & K400   & 43.60 & 27.96 & 26.39 & 14.14 & 22.97 & 12.80     \\
        X3D S    & 13 & K400   & 41.37 & 25.16 & 30.23 & 17.54 & 23.00 & 12.60     \\
        SlowFast & 8  & K400   & 38.82 & 24.22 & \underline{23.34} & \underline{11.70} & 21.39 & 11.57    \\
        SlowFast & 16 & SS V.2 & 39.76 & 23.90 & 30.15 & 17.51 & 21.16 & 11.05    \\ 
        SlowOnly & 8  & K400   & 36.16 & 20.19 & 23.92 & 12.11 & 11.31 & 6.24    \\
        I3D NLN  & 8  & K400   & 38.63 & 24.33 & 24.04 & 12.28 & 20.76 & 11.60     \\
        C2D      & 8  & K400   & \underline{02.63} & \underline{00.97} & 24.33 & 12.39 & \underline{08.85} & \underline{03.92}     \\
        TSM      & 8  & K400   & 45.35 & 23.20 & \textbf{31.81} & \textbf{18.06} & 24.74 & 13.29     \\
        \midrule
        Mean    & -   & -      & 39.51 & 23.98 & 27.37 & 14.90 & 21.01 & 11.38 \\
        STD     & -   & -      & 13.41 & 06.65 & 07.04 & 05.46 & 06.76 & 03.37 \\
        \midrule
        \midrule
        I3D \cite{rodriguez-opazoDORiDiscoveringObject2021} & 8 & Charades & 52.02 & 33.74 & - & - & - & - \\
        I3D \cite{rodriguez-opazoDORiDiscoveringObject2021} & 8 & K400 & - & - & 33.04 & 19.26 & 20.65 & 10.94\\
        \bottomrule
    \end{tabular}%
    }
    \caption{Summary of our experimental results, where $f$ denotes the frame rate used to extract features, while K400 and SS V.2 stand for the Kinetics-400 and Something-Something V.2 datasets, respectively. In the table, the best results for each dataset are indicated in \textbf{bold}, while the worst results are \underline{underlined}.}
    \label{tab:results-table}
    \vspace{-.4cm}
\end{table}

\begin{figure*}[!t]
    \centering
    \includegraphics[width=.3\textwidth]{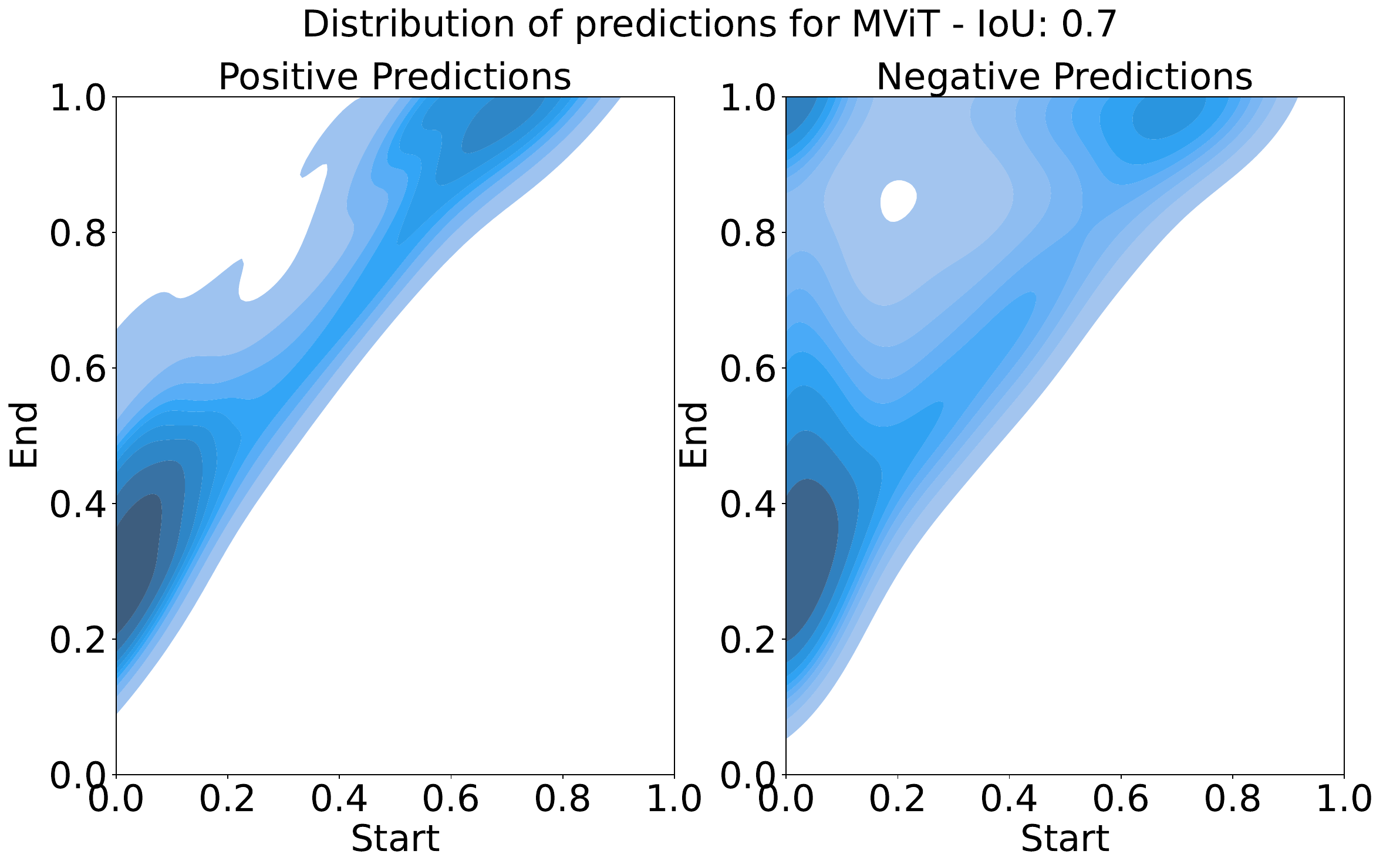} \includegraphics[width=.3\textwidth]{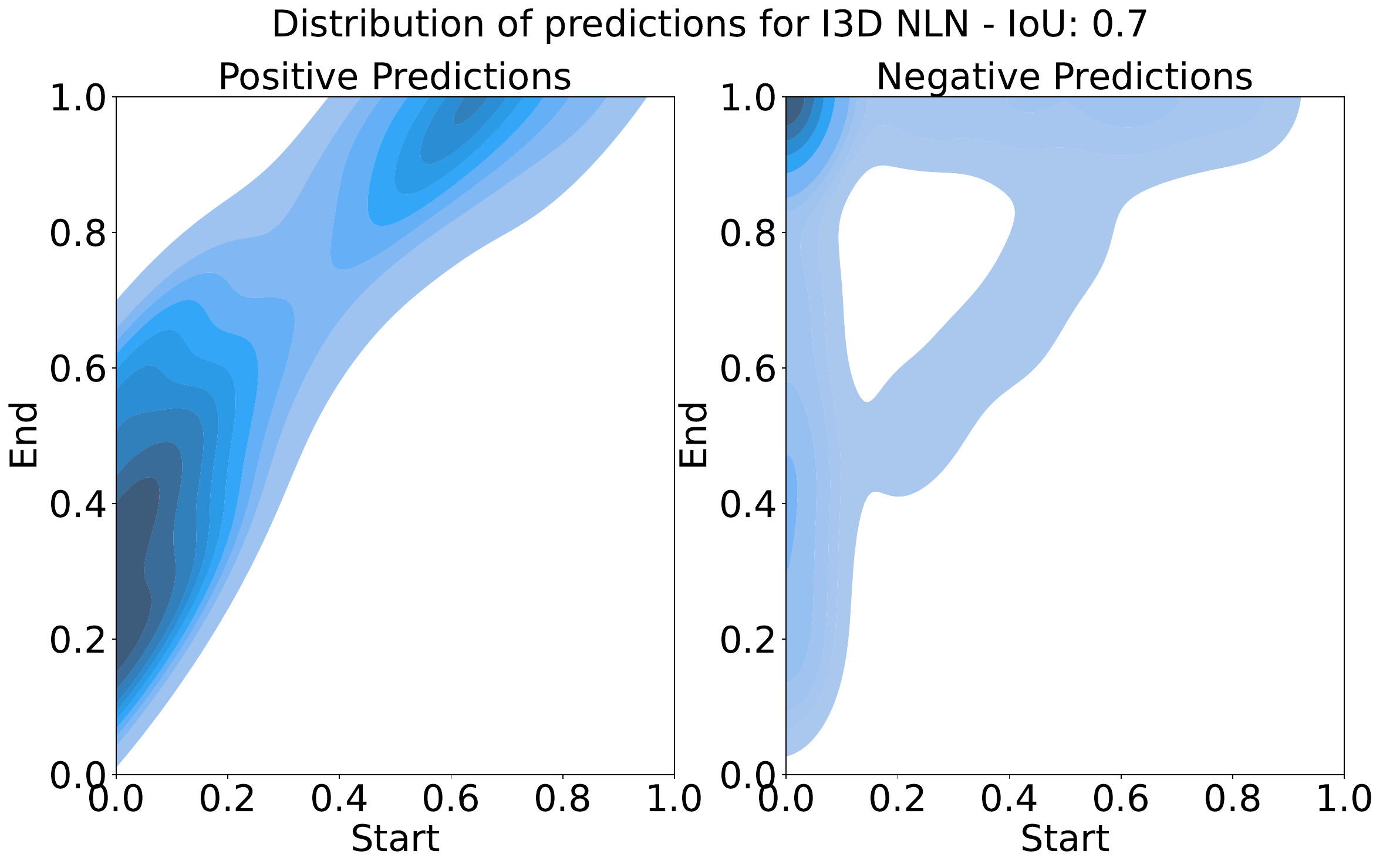} \includegraphics[width=.3\textwidth]{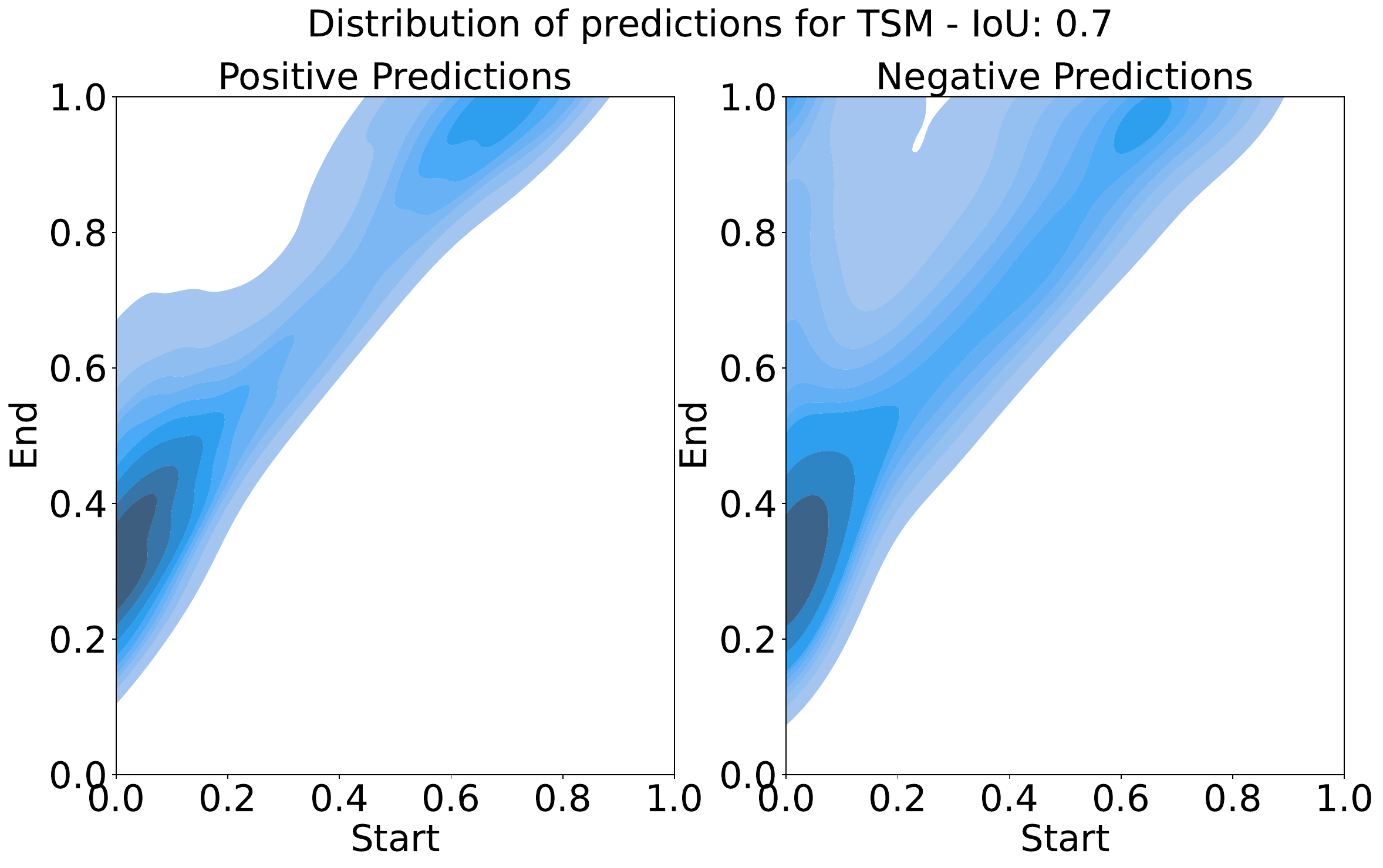}

    \includegraphics[width=.3\textwidth]{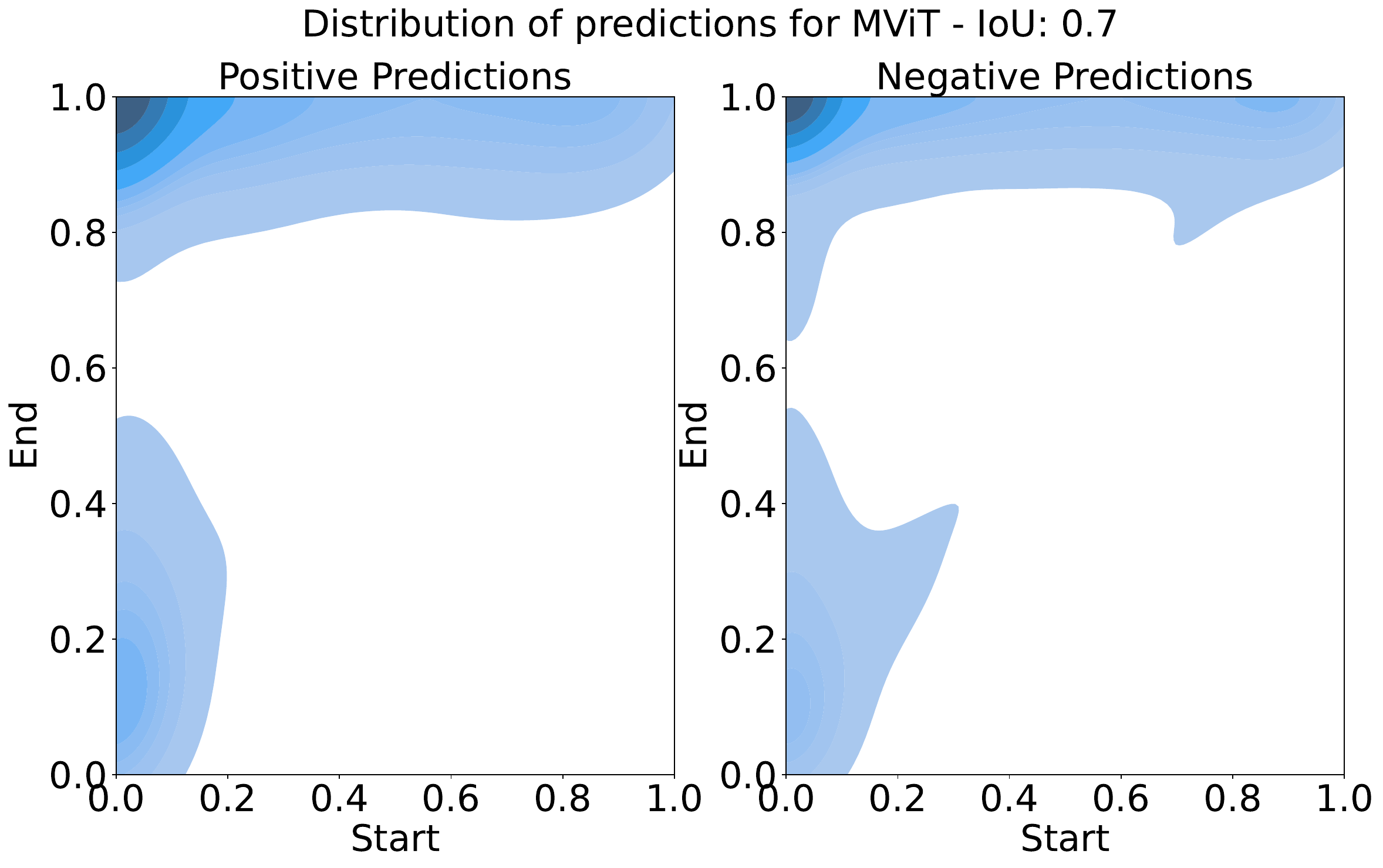} \includegraphics[width=.3\textwidth]{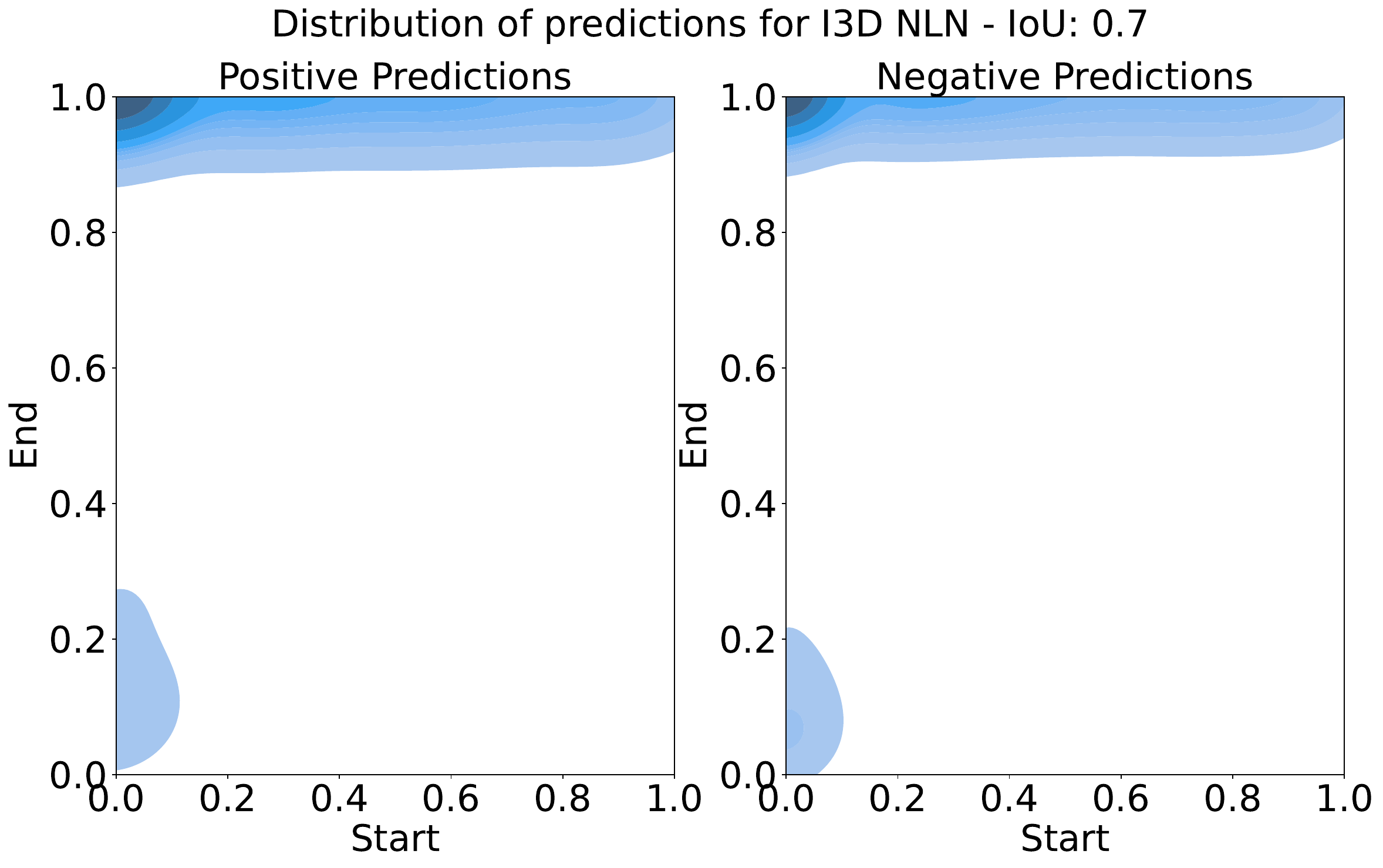} \includegraphics[width=.3\textwidth]{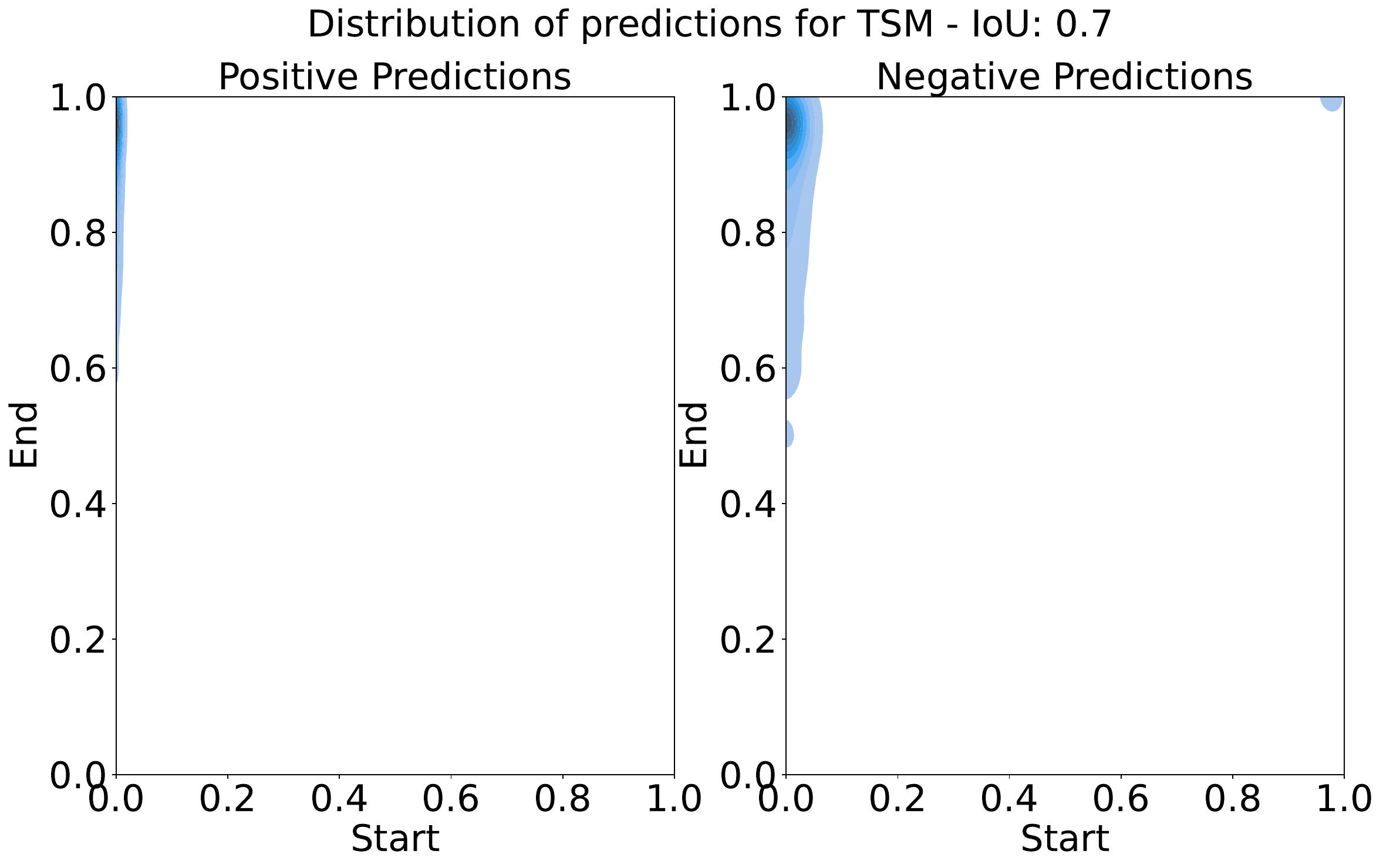}
    
    \caption{The distribution of normalized temporal predictions with extracted features. The x-axis represents the prediction start time, while the y-axis represents the prediction end time. The graph shows normalized predictions for the Charades dataset (top) and the ActivityNet dataset (bottom), using features from MViT (left), I3D NLN (center) and TSM (right).}
    \label{figure:normalized_intervals}
    
\vspace{-.3cm}
\end{figure*}
\begin{figure}[t]
    \centering
    \includegraphics[width=.32\linewidth]{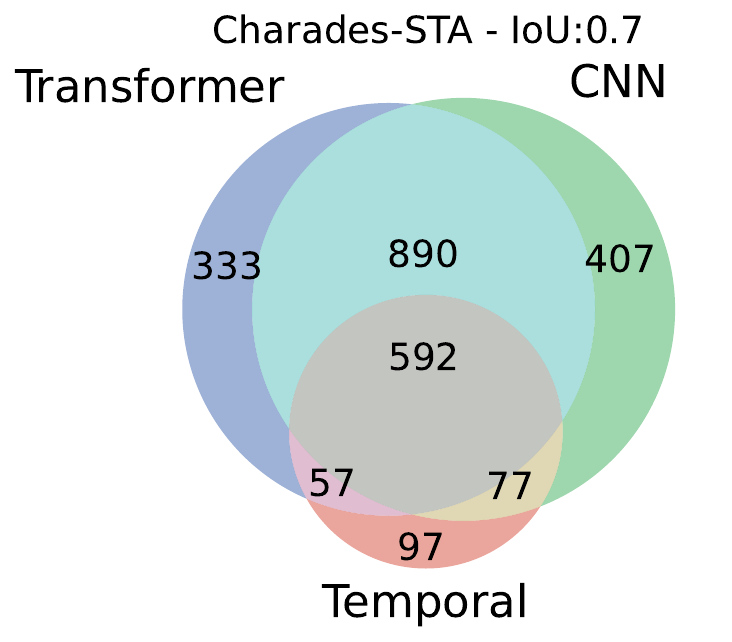}  \includegraphics[width=.32\linewidth]{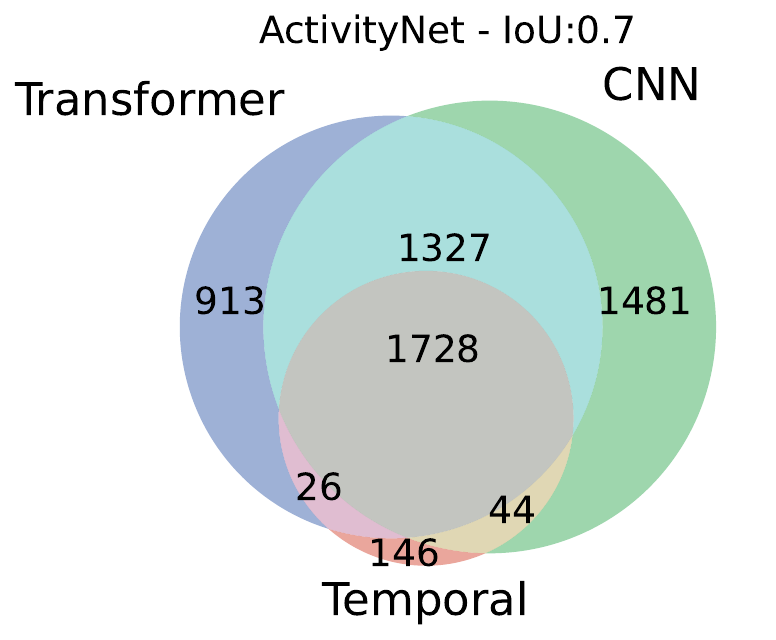}
    \includegraphics[width=.32\linewidth]{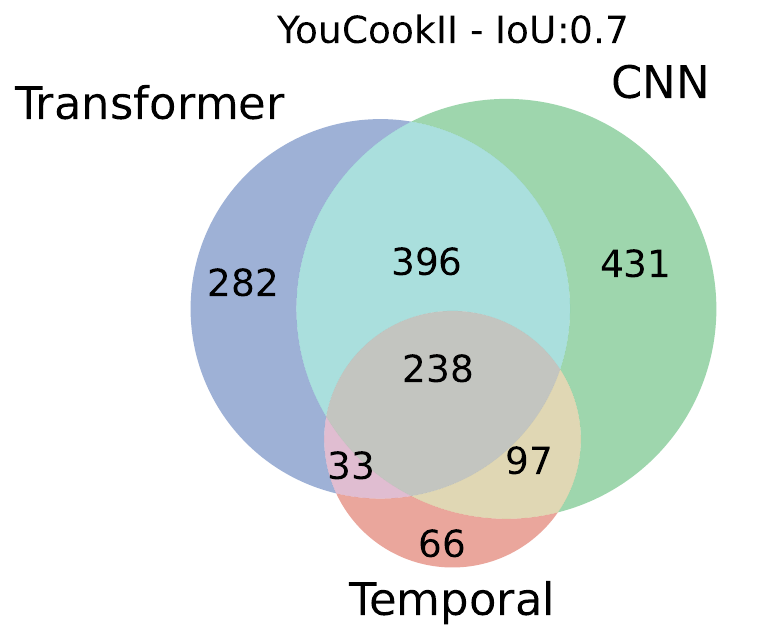}
    \caption{Overlap of the correct predictions, in terms of query-video pairs for $\alpha=0.7$, in Charades-STA (left), ActivityNet (center) and YoucookII (right) benchmarks. We group features based on their nature.}
    \label{figure:venn-diagrams}
    \vspace{-.5cm}
\end{figure}
\paragraph{Quantitative Analysis} Table \ref{tab:results-table} summarizes the results of our experiments, showing the performance of the TMLGA model trained with different features. As can be seen, in the case of the Charades-STA dataset, it is clear that the Transformer-based features give promising results, while our I3D features still perform poorly. This is in contrast to the original results, where the I3D features were only able to deliver $2.45$ points over our results. We attribute this difference to the pre-training dataset, as the original features were pre-trained on Charades, which may have led to overfitting. The proposed features, when pre-trained on the K400 dataset, show competitive performance compared to the original results. This allows for unbiased selection and avoids the need for time-consuming training on the task-specific dataset, thereby accelerating the development of architectures for localization tasks.

On ActivityNet, we note two interesting observations. First, we find that some CNN-based features can outperform more recent Transformer-based models, with differences of up to 7 points in the 0.7 band. We also notice that the overall performance of the features shows less variation, suggesting that the query representation may be particularly relevant. This is consistent with the data showing that ActivityNet has relatively more complex queries in terms of vocabulary size ($748$ vs. $9,744$ tokens) and length ($7.2$ vs. $13.48$ tokens per query).

The results on YouCookII show that this dataset remains the most visually challenging, which we suspect is partly due to the length of the videos, with an average of around 5 minutes. On this dataset, Transformer-based models again achieve the best performance, with results similar to complex architectures such as DORi \cite{rodriguez-opazoDORiDiscoveringObject2021}, which make explicit use of spatial information. 




Overall, our results highlight the importance of careful feature selection in the TVG task, as we see that a simple change can lead to significant performance gains of up to 5 points. Ultimately, we see how selecting features without a thorough understanding of their impact on the dataset can lead to stagnation in development and suboptimal performance. Further research is suggested to investigate the effect of different features and to explore their applicability to other video and speech tasks.




\vspace{-.3cm}
\paragraph{Qualitative Analysis} To visualize the output of the model when given different features, we plot the 0-1 normalized predicted time intervals for each query, similar to \cite{otaniUncoveringHiddenChallenges}. We generate one plot for correct intervals and another for incorrect predictions, based on a given $\alpha$ of 0.7 tIoU band. 

As shown in Figure \ref{figure:normalized_intervals}, we see that in the case of Charades-STA, different features produce two points of correct predictions, represented by the lower-left and upper-right accumulation points. We see that in many cases, the model tends to produce predictions that cover the entire video length, despite the lack of such examples in the training data. We also see the extent to which the predictions vary between features, with clearly different accumulation zones for each case. This indicates that there are significant differences in positive and negative predictions between models on this dataset, suggesting orthogonality between features. For ActivityNet, we observe that the model tends to fall into degenerate solutions in all cases, making predictions for the entire video length. We believe this may be due to the presence of such annotations in the training data. Despite this, similarly to Charades, different accumulation points are evident in the predictions, indicating potential orthogonality between the features.

Finally, as shown in Figure \ref{figure:venn-diagrams}, we plot Venn diagrams to visualize the overlap of correct predictions. We see that there are exclusive predictions based on feature type across the datasets, supporting the visualization of the differences seen in the bias plots. We believe that these results further indicate the presence of orthogonality between the different features, suggesting that the use of different features could be complementary to solving the localization task.



\section{Conclusions}

In this paper, we have conducted a comprehensive analysis of video features in the context of the video grounding task. Our research shows that using features with pre-trained encoders from datasets different from the target application can produce similar results to using a fine-tuned encoder specific to the application dataset. This approach effectively mitigates potential network overfitting and avoids biases arising from assumptions about task proficiency. Our investigation has also revealed exclusive predictive patterns among video features, suggesting the potential for improved predictions through their integration into a unified network. Furthermore, we have shown that varying the features can have a significant impact on localisation, leading to performance improvements comparable to more complex networks that explicitly consider spatial components in their inputs. These results provide valuable insights into how the spectrum of feature diversity and encoder pre-training can be exploited to advance the field of video analysis and improve the field of temporal localisation tasks. Looking ahead, a notable avenue for future work lies in the application of these features within modern architectures. In addition, it is anticipated that the inherent orthogonality of these features will be exploited to build more robust frameworks tailored to the task of temporal localization.


\section*{Acknowledgements}
Ignacio Meza and Felipe Bravo-Marquez were supported by ANID Millennium Science Initiative Program Code ICN17\_002 and the National Center for Artificial Intelligence CENIA FB210017, Basal ANID.

\bibliographystyle{ieee_fullname}
\bibliography{bibliography}

\appendix


\end{document}